\documentclass[letterpaper, 10 pt, journal, twoside]{IEEEtran}
\IEEEoverridecommandlockouts                              

\usepackage{graphicx}
\usepackage{booktabs}
\usepackage{epsfig} 
\usepackage{amsmath} 
\usepackage{amssymb}  
\usepackage{breqn}
\usepackage{cite}
\usepackage{bm}
\usepackage{soul}
\usepackage{multicol, blindtext}
\usepackage[thinlines]{easytable}
\usepackage{float}
\usepackage{siunitx}
\usepackage{xcolor}
\usepackage{soul}
\usepackage{url}
\usepackage{caption}
\usepackage{subcaption}
\usepackage{pdfpages}

\DeclareGraphicsExtensions{.png,.pdf}

\title{The Hydra Hand: A Mode-Switching Underactuated Gripper with Precision and Power Grasping Modes}

\author{Digby Chappell, Fernando Bello, Petar Kormushev, and Nicolas Rojas\vspace{-5pt}%
\thanks{Manuscript received: July 5, 2023; Revised: August 30, 2023; Accepted: September 22, 2023.}
\thanks{This paper was recommended for publication by Editor Hong Liu upon evaluation of the Associate Editor and Reviewers’ comments.}
\thanks{Digby Chappell, Fernando Bello, Petar Kormushev, and Nicolas Rojas are with Imperial College London, London SW7 2BX, UK. Contact:
        {\tt\footnotesize d.chappell19@imperial.ac.uk}. This work was supported in part by the UKRI CDT in AI for Healthcare \texttt{http://ai4health.io} (Grant No. EP/S023283/1).}
\thanks{For the purpose of open access, the author(s) has applied a Creative Commons Attribution (CC BY) license to any Accepted Manuscript version arising.}
\thanks{Digital Object Identifier (DOI): see top of this page.}
}
\begin{document}
\setlength{\belowcaptionskip}{-12.5pt}

\maketitle


\begin{abstract}

Human hands are able to grasp a wide range of object sizes, shapes, and weights, achieved via reshaping and altering their apparent grasping stiffness between compliant power and rigid precision. Achieving similar versatility in robotic hands remains a challenge, which has often been addressed by adding extra controllable degrees of freedom, tactile sensors, or specialised extra grasping hardware, at the cost of control complexity and robustness. We introduce a novel reconfigurable four-fingered two-actuator underactuated gripper---the Hydra Hand---that switches between compliant power and rigid precision grasps using a single motor, while generating grasps via a single hydraulic actuator---exhibiting adaptive grasping between finger pairs, enabling the power grasping of two objects simultaneously. The mode switching mechanism and the hand's kinematics are presented and analysed, and performance is tested on two grasping benchmarks: one focused on rigid objects, and the other on items of clothing. The Hydra Hand is shown to excel at grasping large and irregular objects, and small objects with its respective compliant power and rigid precision configurations. The hand's versatility is then showcased by executing the challenging manipulation task of safely grasping and placing a bunch of grapes, and then plucking a single grape from the bunch.

\end{abstract}


\section{Introduction}\label{sec:intro}
\IEEEPARstart{T}he
grasping versatility of human hands is unparalleled by any other natural or mechanical gripper. 
Human hands are the result of a complex system of motor and sensory pathways that enable humans to switch at ease between precision and power grasps, and offering compliance and adaptability to varying shapes, sizes, and weights \cite{Sobinov2021TheDexterity}.
To achieve this in a robotic gripper is extremely challenging, with grasped object variability, unstructured grasping environments, multi-contact sensing difficulties, and high occlusion manipulation all contributing towards complexity.
Achieving anywhere near human-level versatility with one gripper and a reduced number of actuators is non-trivial, and indeed finding an optimal method of doing so remains an open area in robot manipulation research.

\begin{figure}[!t]
    \centering
    \includegraphics[width=0.85\columnwidth]{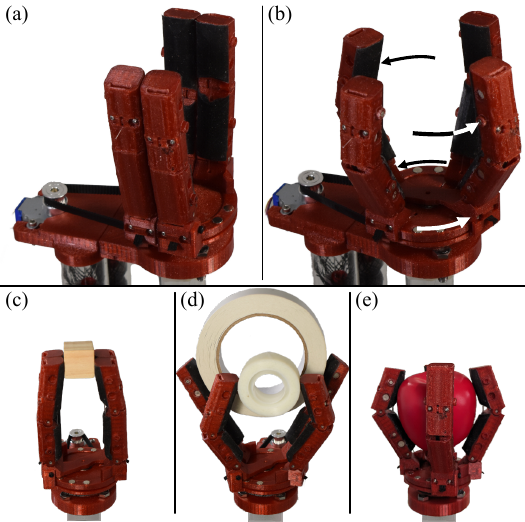}
    \vspace{-5pt}
    \caption{The Hydra Hand, capable of performing rigid precision and compliant power grasps with a single hydraulic grasp actuator, where mode-switching is achieved via a rotating palm. (a) Two-fingered precision grasping mode. (b) Four-fingered spherical grasping mode. (c) Rigid precision palmar pinch grip. (d) Cylindrical power grip with adaptability between pairs of fingers. (e) Spherical power grip with individual finger compliance. }
    \label{fig:top_level}
\end{figure}

Many works have approached the grasping versatility problem from a control and sensory perspective. Often, tactile sensors~\cite{VanHoof2015LearningFeatures} and vision systems~\cite{OpenAI2019SolvingHand} are used, or computational models of the interaction between the gripper and known objects are developed~\cite{Hang2021ManipulationManipulation}. These studies display impressive dexterity with known objects, but are limited in robustness and how well they generalise to grasping unknown and varied objects. Furthermore, grasping in unstructured environments is difficult from a sensory perspective, where high occlusion and multiple contact points with objects increases complexity significantly.

\begin{figure*}[t!]
    \centering
    \vspace{8pt}
    \includegraphics[width=0.9\textwidth]{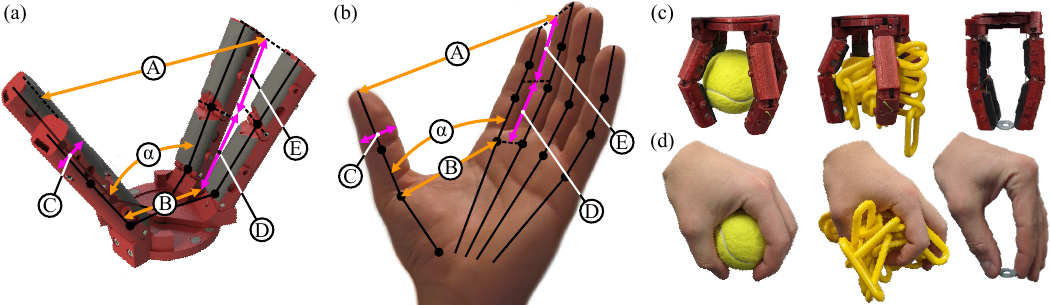}
    \vspace{-5pt}
    \caption{A comparison between the dimensions and grasping actions of the Hydra Hand and a human hand. Key dimensions of the Hydra Hand (a) are modelled on corresponding dimensions of a human hand (b). The Hydra Hand (c) is able to imitate human grasps (d), specifically those involving form-closure and palmar pinch grips.}
    \label{fig:mode_switching}
\end{figure*}

A related approach is to increase the number of degrees of freedom (DOFs) in the gripper such that an increased number of grasps can be achieved. Many anthropomorphic hands utilise this strategy, such as the ILDA hand~\cite{Kim2021IntegratedHand} and the Robonaut Hand~\cite{Lovchik1999RobonautSpace}. Some non-anthropomorphic hands, such as the farmHand~\cite{Ruotolo2021FromGripper} and the Yale Model W~\cite{Bircher2021ComplexCaging} also follow this trend, with targeted DOFs for desired grasps. However, increasing the number of controllable DOFs increases control complexity, and ensuring constraints such as grasp success and safety are satisfied can be difficult with a larger action space.

Rather than increasing gripper complexity, an alternative method is to optimise a small number of DOFs to achieve a wider range of manipulation tasks.
These reduced-complexity grippers are usually designed for in-hand manipulation, with specialised DOFs for reorienting and translating grasped objects. Common design features include variable friction surfaces~\cite{Spiers2018Variable-FrictionSliding, Lu2020AnGrippers} and the ability to reconfigure the fingers of the gripper~\cite{Townsend2000TheAssembly, Rojas2016TheManipulation, Lu2021SystematicGripper, Kang2021ModelingObjects}. These works show robust in-hand manipulation results with a range of grasped objects, however these grasped objects are often limited to larger diameters with compliant grasps.
Achieving compliance with a small number of degrees of freedom requires underactuation, where a small number of control inputs actuate multiple joints indirectly. The most common technique of achieving compliance in this way is to have a single tendon routed through multiple joints, as in the RUTH gripper~\cite{Lu2021SystematicGripper} and the OLYMPIC hand~\cite{Liow2020OLYMPIC:Mechanisms}.
Reducing the number of actuators even further is often achieved with a differential mechanism in the palm of the hand that allows non-contacting fingers to continue flexing when one becomes blocked. Examples of these are often present in prostheses, where mechanical and control complexity must be minimised, such as the Hannes hand~\cite{Laffranchi2020TheHand}, which uses a series of balanced pulleys, or the soft prosthesis presented in \cite{Gu2021AFeedback}, which is pneumatically actuated such that actuation pressure is distributed across the finger. However, these hands are ill-suited to precision tasks, being designed primarily for power grasps.

Reduced-order grippers designed for both compliant and precise grasps are much rarer. The Meso gripper uses a single actuator to achieve power and pinch grasps with a two-fingers~\cite{Bai2017KinematicMeso-gripper}, however the Meso gripper's contact region is limited to its fingertips, while the remainder of each finger is a fully actuated mechanism. Adaptability in this case is achieved via a differential pulley in the palm, and precision grasping relies on zero surface contact before the grasp --- something that is difficult to achieve when grasping small objects. The Velo gripper~\cite{Ciocarlie2014TheGrasps}, also a two-fingered single-actuator gripper, uses underactuation to its advantage in precision grasping. Each finger of the Velo gripper is balanced with an extension tendon that causes the distal phalanges to form a parallel grasp by default, meaning that an enveloping grasp---a necessity for heavier objects---only occurs when an object is actively blocking the proximal phalanges. A significant drawback of both the Meso and the Velo grippers is that they are two-fingered, restricting grasping to a single axis and reducing their form-closure ability. As such, the use of minimal actuators for multi-fingered precision and compliance is a potentially fruitful avenue of research that is presently unexplored.

In this work, we present the Hydra Hand {(shown in Fig.~{\ref{fig:top_level}})}: a four-fingered gripper that is capable of switching between power and precision modes, and is operated by a single grasping actuator and a single switching actuator. By utilising McKibben muscles to drive fingers in pairs, pressure generated from a single input is shared across four fingers, allowing compliance and independent motion of finger pairs during power grasping. Deliberately misaligned flexion axes of adjacent fingers allow the hand to switch modes from four underactuated fingers for compliant grasping to two rigid fingers for precision grasping with a palm rotation mechanism.
We test the gripper on a range of household benchmarking objects and textiles, and find that it excels at both power and precision grasps, and can adapt its grip to accommodate objects of varying size and shape.
We demonstrate the versatility of the gripper in a challenging manipulation scenario: grasping a bunch of grapes, then picking a single grape from the bunch.


\section{Design}\label{sec:design}
The Hydra Hand is designed to perform both rigid precision and compliant power grasps. The targeted precision grasp is a two-fingered pinch, which can take the form of a palmar pinch or an inferior pincer~\cite{Feix2016TheTypes}. For power grasps, reshaping the hand and caging objects in three dimensions becomes more important, so both cylindrical and spherical power multi-finger grasps are targeted. Furthermore, the shape of power-grasped objects can be irregular, therefore a level of adaptability between compliant fingers is necessary to accommodate for this. The design of the Hydra Hand is split into three subsections: the mode-switching mechanism, the kinematics of the gripper, and finger actuation.

\begin{figure}[t!]
    \centering
    \vspace{8pt}
    \includegraphics[width=0.95\columnwidth]{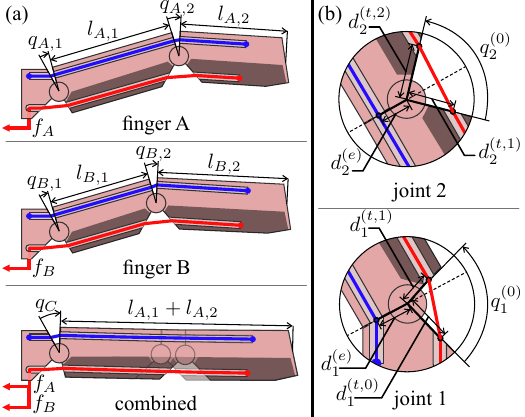}
    \vspace{-3pt}
    \caption{Schematic diagram of the fingers and joints of the Hydra Hand. (a) Individual $A$ and $B$ fingers and the combined $A+B$ fingers. (b) Joints of each finger - note that all fingers are designed with the same tendon routing offsets and initial angular offsets. Flexion tendons are shown in red on the palmar side of each finger. Extension elastic cords are shown in blue on the dorsal side of each finger. }
    \label{fig:kinematics}
\end{figure}

The dimensions of the Hydra Hand are modelled on those of an adult male's hand (see Fig.~\ref{fig:mode_switching}). Each finger is approximately $95$~mm long and has a width and thickness of $20$~mm, falling in the range of the average index finger length and average thumb width and thickness. The palm of the Hydra Hand has a diameter of $60$~mm, which is designed to match the average distance from the base of the thumb to the base of the index finger on the palmar surface of a human hand. At rest the fingers are at an angle of $30^{\circ}$ to each other, giving a fingertip-to-fingertip distance of approximately $140$~mm, mimicking the angle and fingertip-to-fingertip distance of a human hand when reaching to grasp a medium-sized object. The hand is comprised of four fingers: two type $A$ fingers, and two type $B$ fingers {(shown in Fig.~{\ref{fig:kinematics}})}; the length of the proximal and distal phalanges of the type $A$ fingers are $50$~mm and $45$~mm, respectively, and of the type $B$ fingers are $40$~mm and $55$~mm, respectively.

\subsection{Mode-Switching}
The type $A$ fingers of the Hydra Hand are mounted on a static palm section, while the type $B$ fingers are mounted on a rotating palm section. The rotating palm section is actuated by a stepper motor, and allows the hand to reconfigure and reshape to achieve a range of grasps between cylindrical and spherical grips. When the rotating palm reaches its limit, adjacent fingers magnetically lock together and produce a precision gripper. The base of each finger is deliberately mounted at a lateral offset from the sagittal plane of the hand, meaning the flexion planes of all four fingers are parallel when the hand is in precision mode. A similar rotating palm is utilised in the Yale Model Q, but is primarily used to perform in-hand rotation of objects, rather than to expand the available grasps of the gripper~\cite{Ma2014AnManipulation}.

Unlike a human finger, that can flex its proximal and distal joints independently, the Hydra Hand is underactuated, with a single tendon controlling both finger joints. Because it is not possible to purposely position the fingertips such that a fingernail feature on the dorsal side of the finger could be used for a tip pinch grip, the targeted precision grasps are the palmar pinch and inferior pincer grasps. To aide this, a fingernail feature is present on the distal face of the fingertip, allowing the palmar pinch grasp to also serve as a tip pinch.
\begin{figure}[t!]
    \centering
    \vspace{8pt}
    \includegraphics[width=0.875\columnwidth]{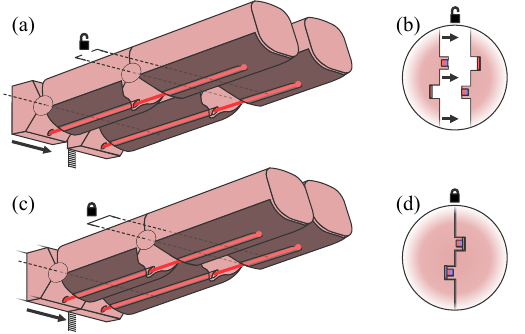}
    \vspace{-4pt}
    \caption{The finger locking mechanism of the Hydra Hand. (a) and (c), Flexion axes of the distal joints of each finger are misaligned. (b) and (d) Locating magnetic features embedded in each finger prevent relative {in-plane} rotation and translation of laterally adjacent links.}
    \label{fig:finger_locking}
\end{figure}
The motion group of finger $A$ can be described by sequential rotations around joint axes $a_1$ and $a_2$~\cite{Rojas2016GrossManipulation}:
\begin{equation}
    \mathcal{S}_A = \{\bm{R}(A_1, a_1)\}\cdot\{\bm{R}(A_2, a_2)\},
\end{equation}
and similarly for finger $B$, with joint axes $b_1$ and $b_2$: 
\begin{equation}
    \mathcal{S}_B = \{\bm{R}(B_1, b_1)\}\cdot\{\bm{R}(B_2, b_2)\}.
\end{equation}
When the rotating palm section reaches its limit, the finger pairs are aligned and passively locked together with raised features and magnetic contacts (see Fig. \ref{fig:finger_locking}) to act as a single combined finger for two-fingered grasping. To switch from compliant fingers to rigid fingers for precision grasping, the distal flexion axes of the fingers are misaligned. The raised features prevent the relative {in-plane} motion of the adjacent distal links, and therefore constrain the flexion of the distal joints to be zero. The only joint that can therefore be actuated is the combined proximal joint, producing a two-fingered rigid gripper that can perform precision grasps. When finger $A$ and $B$ are combined, the combined finger motion group is the intersection of the two component motion groups, $\mathcal{S}_{A+B}=\mathcal{S}_A\cap\mathcal{S}_B$:
\begin{align}
    \mathcal{S}_{A+B} &= (\{\bm{R}(A_1, a_1)\}\cdot\{\bm{R}(A_2, a_2)\})\cap\nonumber\\&(\{\bm{R}(B_1, b_1)\}\cdot\{\bm{R}(B_2, b_2)\}).
\end{align}
The laterally adjacent links of $A$ and $B$ are rigidly connected, and axes $a_1$ and $b_1$ are aligned, therefore $\{\bm{R}(A_1, a_1)\}=\{\bm{R}(B_1, b_1)\}$ and $\mathcal{S}_{A+B} = \{\bm{R}(A_1, a_1)\} (\{\bm{R}(A_2, a_2)\}\cap\{\bm{R}(B_2, b_2)\})$. Since $a_2$ and $b_2$ are parallel but offset, $\{\bm{R}(A_2, a_2)\}\cap\{\bm{R}(B_2, b_2)\}=\{\bm{I}\}$ is identity displacement. The combined motion group of displacement is therefore:
\begin{equation}
    \mathcal{S}_{A+B} = \{\bm{R}(A_1, a_1)\}.
\end{equation}
{In the direction perpendicular to the flexion plane of the fingers, the fingers are only held in contact by the attractive forces of the magnets; these forces are not large and allow the fingers to be unlocked by reversing the direction of the rotating palm unit.} This simple mechanism allows the hand to switch from compliant power grasps to rigid precision grasps {and vice versa} without any dedicated hardware changes.
\begin{figure}[t!]
    \centering
    \vspace{8pt}
    \includegraphics[width=0.85\columnwidth]{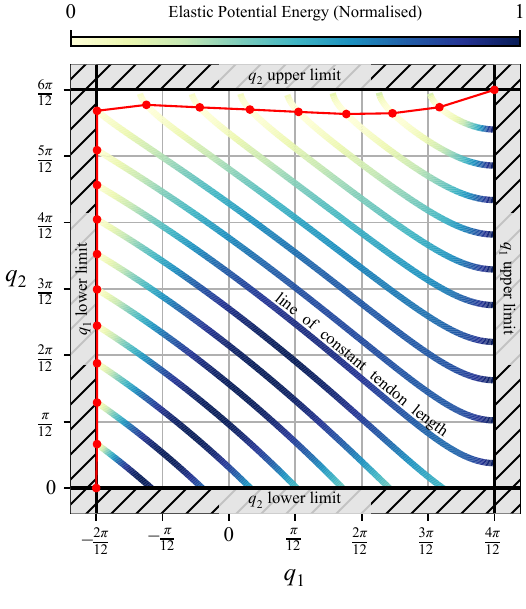}
    \vspace{-12pt}
    \caption{Joint space of an individual finger, with joint limits and lines of constant tendon length shown. Lines of constant tendon length are coloured according to the amount of stored elastic potential energy in the elastic extension cord, with lighter showing less energy, and darker showing more energy (normalised per line). {The minimum-energy unhindered joint-space trajectory of a finger is shown from rest ($-\frac{\pi}{6}$, $0$) to maximum flexion ($\frac{\pi}{3}$, $\frac{\pi}{2}$) in red.}}
    \label{fig:joint_space}
\end{figure}
\subsection{Gripper Kinematics}
During power grasping, the fingers of the hand are underactuated, with a single tendon driving two flexion joints of each finger, and elastic cord passively driving the extension of both joints (see Fig. \ref{fig:kinematics} (a)). The tendon is routed inside each finger such that it directly acts on each joint, rather than being routed around a joint pulley. The underactuation of each finger enables the hand to be compliant---if the proximal joint gets blocked then the distal joint is free to continue closing. The retractable tendon length $L_{f, j}^{(t)}$ of joint $j$ of finger $f$ can be found using the cosine rule with joint angle $q_{f,j}$ and, with reference to Fig.~\ref{fig:kinematics}(b), tendon routing points at perpendicular offsets of $d_{j,l}^{(t)}$ on link $l$ and $d_{j,l-1}^{(t)}$ on the preceding link, with an offset angle of $q_{j}^{(0)}$:
\begin{equation}
    L_{f, j}^{(t)} = \sqrt{d_{j,l}^{(t)^2} + d_{j,l-1}^{(t)^2} - 2d_{j,l}^{(t)}d_{j,l-1}^{(t)}\cos{(q_{j}^{(0)} - q_{f,j})}},
\end{equation}
with a similar formulation for the extendable cord length $L_{f, j}^{(e)}$, noting that (see Fig. \ref{fig:kinematics}(b)) the perpendicular offset of the routing point of the elastic cord is identical for both links at $d_{j}^{(e)}$, and there is zero initial angular offset:
\begin{equation}
    L_{f, j}^{(e)} = d_{j}^{(e)}\sqrt{2(1 - \cos{(q_{f,j})})}.
\end{equation}
The perpendicular distance of the tendon to the joint, $L_{f, j}^{(t, \perp)}$ can be found using the angle $\alpha_{j,l}^{(t)}$ that $d_{j,l}^{(t)}$ makes with the tendon, $L_{f, j}^{(t, \perp)} = d_{j,l}^{(t)}\sin(\alpha_{j,l}^{(t)})$, where $\sin(\alpha_{j,l}^{(t)})$ can be found using the sine rule:
\begin{equation}
    \sin(\alpha_{j,l}^{(t)}) = \frac{d_{j,l}^{(t)}}{L_{f, j}^{(t)}}\sin(q_{j}^{(0)} - q_{f,j}).
\end{equation}
This results in a perpendicular tendon distance of
\begin{equation}
    L_{f, j}^{(t, \perp)} = \frac{d_{j,l}^{(t)^2}}{L_{f, j}^{(t)}}\sin(q_{j}^{(0)} - q_{f,j}),
\end{equation}
and a similar result can be found for the perpendicular distance of the elastic cord $L_{f, j}^{(e, \perp)}$:
\begin{equation}
    L_{f, j}^{(e, \perp)} = \frac{d_{j,l}^{(e)^2}}{L_{f, j}^{(e)}}\sin(q_{f,j}).
\end{equation}

When the tendon is stationary, the fingers will move to the position of minimum energy. Neglecting gravitational potential energy, the elastic potential energy stored in the elastic extension cord mounted on the dorsal side of the finger is the driving factor in the stable joint-space configurations of the finger. The change in elastic potential energy from the initial position of the finger ($L_{f, 1}^{(e)}=L_{f, 2}^{(e)}=0$) can be computed given spring constant $k_f^{(e)}$:
\begin{equation}
    E_f^{(e)} = \frac{1}{2}k_f^{(e)}(L_{f, 1}^{(e)} + L_{f, 2}^{(e)})^2.
\end{equation}

Fig. \ref{fig:joint_space} shows the joint space of a finger of the Hydra Hand, with lines of constant tendon length shown, coloured according to normalised elastic potential energy. As seen, the minimum-energy joint-space trajectory of the finger is such that $q_{f,1}$ is minimised until $q_{f,2}$ reaches its joint limit---this is useful for caging objects---however, also seen is another region of minimum-energy is when $q_{f,2}$ is minimised, indicating that bistable grasping behaviour could be achieved with small design alterations. Bistable mechanisms have been used in literature to produce secure, low actuation energy grasps~\cite{Stavenuiter2017ACompliance, Mouaze2022BistableObjects}, and there is scope for future work to explore this for the Hydra Hand. Many underactuated grippers balance the extension springs such that the proximal joints are energetically favourable to actuate, meaning a precision/parallel grasp is the default behaviour of the gripper, and the distal joints do not flex until the proximal phalanges are physically blocked by a grasped object~\cite{Ciocarlie2014TheGrasps}. This form of compliance is favourable when grasping rigid objects, but may not be suitable for delicate or articulated objects, where caging the object via form closure may be a safer solution.

\subsection{Hydraulic Actuation and Adaptability}
\begin{figure}[t!]
    \centering
    \vspace{8pt}
    \includegraphics[width=0.85\columnwidth]{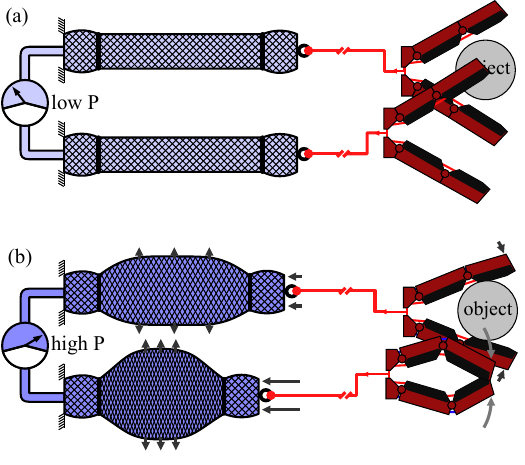}
    \vspace{-8pt}
    \caption{Tendon actuation via McKibben muscles. (a) Relaxed state. (b) Contracted state. When the motion of one tendon is blocked, pressure is distributed to the other actuating muscle. This allows independent tendon motion. }
    \label{fig:muscles}
\end{figure}
During operation in power grasping mode the fingers of the gripper should move with a degree of independence in order to accommodate for irregular objects of unknown size and shape. This is difficult with a single actuator, with existing hands utilising pulley balancing systems~\cite{Ko2020AOptimization, Laffranchi2020TheHand, Ciocarlie2014TheGrasps, Catalano2014AdaptiveSoftHand, Dollar2007SimpleEnvironments} and other differentials~\cite{Lu2021AnRejection, Lu2021MechanicalGrasp, Xu2015AMechanism} to achieve this. A simple alternative is to use hydraulic actuation, whereby pressure can be equally shared across the fingers; if one finger makes contact with an object, pressure continues to actuate the others. This may be preferable in some scenarios over a tendon-based differential, where friction builds exponentially with tendon length in the absence of suitable bearings. However, with a tendon-driven hand achieving this requires a method of converting pressure to tendon motion. To achieve this, we utilise McKibben muscles \cite{Gaylord1955FluidDevice} to convert fluid pressure to linear motion. Shown in Fig. \ref{fig:muscles}, two muscles are used --- one for each pair of fingers. {When the gripper is in precision mode, one finger from each pair makes up the combined finger, ensuring that the two combined fingers flex as one fully actuated gripper. This is beneficial when grasping small flat objects; if the object is laterally offset then the closest finger will push the object towards the centreline of the gripper, and the opposing finger will not swing through prematurely. In power grasping mode, it is possible that irregularly shaped objects can block the motion of a finger pair, leaving one finger without contact. However, given that the object is graspable, it will move towards the centreline of the gripper until a rigid grasp with a minimum of three points of contact have been achieved.} Water was chosen as a hydraulic fluid due {to} its low viscosity and incompressibility under normal conditions, effectively converting muscle volume change directly to tendon contraction.

\section{Performance Evaluation}\label{sec:evaluation}
\begin{figure}[t!]
    \centering
    \vspace{8pt}
    \includegraphics[width=0.8\columnwidth]{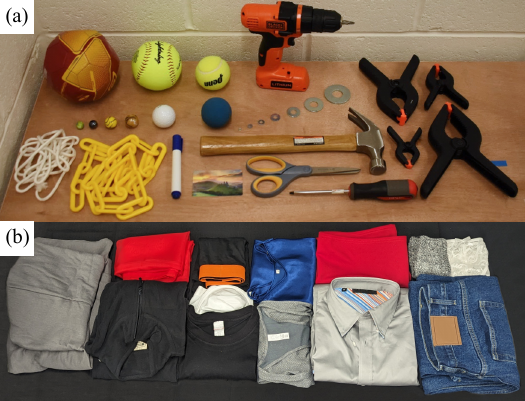}
    \vspace{-5pt}
    \caption{Objects used for the YCB Gripper Assessment Benchmark (a) and Soft Object Gripper Benchmarks (b).}
    \label{fig:benchmark_objects}
\end{figure}

The Hydra Hand was evaluated on the YCB Gripper Assessment Benchmark~\cite{Calli2015BenchmarkingSet} in each of its three grasping configurations: precision pinch grasp, cylindrical power grasp, and spherical power grasp. The benchmark consists of $28$ objects grasped a total of $136$ times - each rigid object is grasped in a series of positions selected from the following: $\mathcal{O}$ (aligned with the gripper), $\Delta x$, $\Delta y$, and $\Delta z$ (offset by $1$~cm in the $x$, $y$, and $z$ directions, respectively). This was repeated for each grasping mode, giving a total of $408$ attempted grasping actions. The objects included in the benchmark can be seen in Fig. \ref{fig:benchmark_objects} (a).

To quantify the Hydra Hand's ability to grasp deformable objects, three of the Soft Object Gripper Benchmarks~\cite{Clark2023HouseholdManipulation} were performed: `A) flat edge grasp success and drag placement accuracy', `C) crumpled object encapsulation', and `D) flat non-boundary grasp success'. These were chosen to capture both precise and power grasping ability. Benchmark `B) edge grasp resilience' was not performed, as it is a measure of grasping force, rather than the ability to grasp. The items of clothing used in this benchmark are shown in Fig. \ref{fig:benchmark_objects} (b), and $330$ attempted grasps were performed in total.

\begin{figure*}[t!]
    \centering
    \vspace{8pt}
    \includegraphics[width=0.95\textwidth]{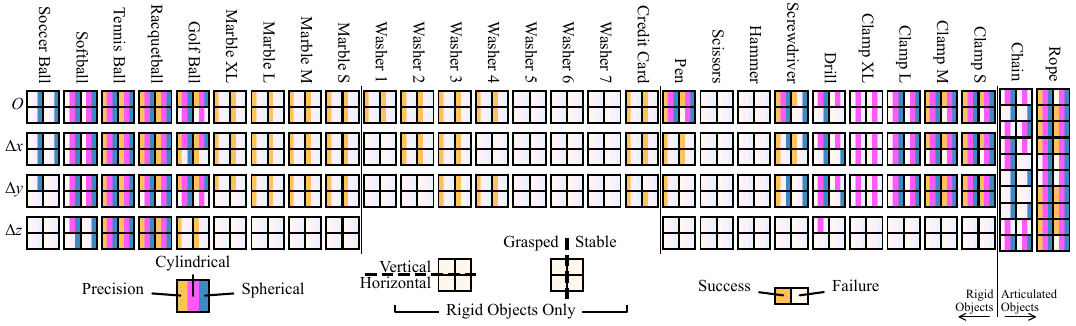}
    \vspace{-8pt}
    \caption{Results of the YCB Gripper Assessment Benchmark. For rigid objects, in each $2\times2$ square, cells are coloured dark or light according to respective success or failure according to the following criteria: top left if a successful grasp is achieved in the vertical orientation, top right if the object is static for $3$ seconds after being grasped from above, bottom left if the object remains held while rotated to horizontal, and bottom right if the object is completely static for $3$ seconds after being rotated. This was repeated at an origin point $\mathcal{O}$, and at $1$~cm offsets $\Delta x$, $\Delta y$, and $\Delta z$. For articulated objects, there are $20$ cells showing attempted grasp success or failure, coloured dark or light accordingly. Each cell is divided according to grasp mode used: precision, cylindrical, and spherical modes are shown in orange, magenta, and blue, respectively.}
    \label{fig:ycb_results}
\end{figure*}

\section{Results \& Discussion}\label{sec:results}
\begin{figure}[t]
  \centering
  \includegraphics[width=0.9\columnwidth]{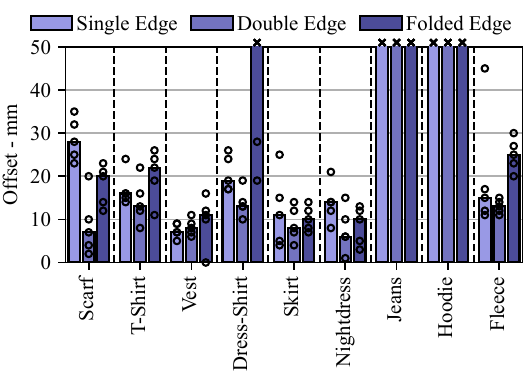}
  \vspace{-4pt}
  \caption{Results of the Soft Object Gripper Benchmark A. Items of clothing are grasped and dragged $500$~mm horizontally, and placement error is measured. Results for single, double, and folded edges are shown. Bars show median result, $\circ$ and $\times$ marks show successful and failed attempts, respectfully.}
  \label{fig:soft_results_a}
\end{figure}

\begin{figure*}[t]
  \vspace{0pt}
  \centering
  \includegraphics[width=0.85\textwidth]{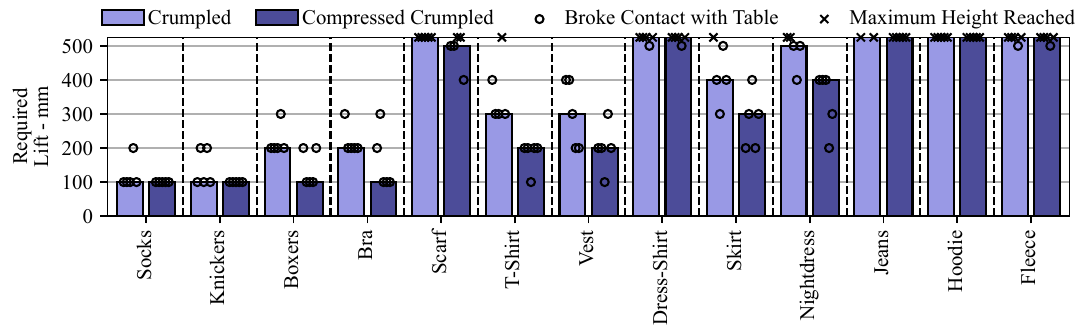}
  \vspace{-3pt}
  \caption{Results of the Soft Object Gripper Benchmark C: items of clothing are crumpled, then grasped and lifted until contact with the table is completely broken. Lift height required to break contact with the table (of successful grasps) is shown; Bars show median result, $\circ$ and $\times$ marks show lifts that did and did not break contact with the table, respectively. Note that the Hydra Hand failed to grasp the jeans $3/5$ times in the crumpled state.}
  \label{fig:soft_results_c}
\end{figure*}

The results of the YCB Gripper Assessment Benchmark are shown in Fig. \ref{fig:ycb_results}. As expected, the two compliant power grasping modes fail to grasp small and flat objects, while the precision grasping mode fails to grasp irregularly shaped tool objects. Individually, the precise, cylindrical, and spherical grasping modes achieve scores of $175/404$, $132.5/404$, and $140/404$, respectively. Together, the combined score of the Hydra Hand is $257.5/404$ (where the result for each object is taken as the best from the three grasping mode results) --- comparable to the Yale Model B gripper~\cite{Backus2018AVehicles}, also utilising a single grasping actuator. Fig. \ref{fig:soft_results_a}, Fig. \ref{fig:soft_results_c}, and Table \ref{fig:soft_results_d} show the results of the Soft Object Gripper Benchmarks A, C, and D, respectively. Benchmark A, where the gripper must grasp cloth edges and drag the item of clothing $500$~mm horizontally, and Benchmark D, where each garment was grasped from a central featureless region, were completed using the precision grasping mode. Benchmark C, where crumpled items of clothing are grasped then raised until completely lifted from the table surface, was completed using the spherical power grasping mode.

The precision mode was successful in grasping the majority of small and flat objects in the YCB Gripper Assessment Benchmark (see Fig. \ref{fig:ycb_results}) and was able to grasp all single, double, and folded edges in the Soft Object Gripper Benchmark A (see Fig. \ref{fig:soft_results_a}), as well as successfully grasping nearly all clothes at a central featureless region in Soft Object Gripper Benchmark D (see Table. \ref{fig:soft_results_d}). However, some failure cases were observed. The first case was when the object was too small in height, such as the smallest three washers of the YCB Gripper Benchmark. The fingertips of the gripper have a `fingernail'-style feature, which is a rigid lip designed to guide objects upwards into the hand by a small amount, however this fingernail has a finite thickness and it was not possible to exploit this feature when grasping small, flat objects. The second case was when grasping heavier objects and stiffer materials. Because only the proximal joint is active during precision mode, the achievable grasping force of the gripper is reduced compared to the compliant modes, and objects located at the distally will require more actuation force. For example, grasping the largest washer at any offset, grasping central regions of the denim jeans, and dragging the grasped edges of the hoodie and jeans were not possible with the Hydra Hand. In practice, grasping strategies to bring the objects closer to the palm can be used to negate this. In precision mode, adaptive grasping between pairs of fingers is removed to produce a repeatable central grasp. This was especially important when grasping edges of clothing, where it is difficult to predict when one finger will contact the garment. Had adaptive grasping been utilised, premature contact would have allowed the non-contacting finger to move past the target grasping point. {Due to the mechanical tolerances required for the raised locking features to easily connect and disconnect during locking and unlocking, a slight amount of flexion of the distal finger joints was observed in precision mode. This slight flexion was negligible and did not impair precision grasping.}

\setlength{\tabcolsep}{0.3em} 
{\renewcommand{\arraystretch}{1.6}
\begin{table}[t]
    \centering
    \begin{tabular}{|c|c|c|c|c|c|c|c|c|c|c|c|c|}
        \hline
        \rotatebox[origin=c]{90}{Socks} & \rotatebox[origin=c]{90}{Knickers} & \rotatebox[origin=c]{90}{Boxers} & \rotatebox[origin=c]{90}{Bra} & \rotatebox[origin=c]{90}{Scarf} & \rotatebox[origin=c]{90}{T-Shirt} & \rotatebox[origin=c]{90}{Vest} & \rotatebox[origin=c]{90}{Dress-Shirt} & \rotatebox[origin=c]{90}{Skirt} & \rotatebox[origin=c]{90}{Nightdress} & \rotatebox[origin=c]{90}{Jeans} & \rotatebox[origin=c]{90}{Hoodie} & \rotatebox[origin=c]{90}{Fleece}
        \\
        \hline
         1.0 & 1.0 & 1.0 & 1.0 & 1.0 & 0.8 & 1.0 & 0.8 & 1.0 & 1.0 & 0.0 & 1.0 & 1.0  \\
        \hline
    \end{tabular}
    \caption{Results of the Soft Object Gripper Benchmark D. Items of clothing are grasped at a flat, non-boundary region, and grasp success rate is recorded.}
  \label{fig:soft_results_d}
\end{table}

The cylindrical mode was only used in the YCB Gripper Assessment Benchmark (see Fig. \ref{fig:ycb_results}), and excelled at grasping irregularly shaped tool objects (the clamps), and mid-sized spherical objects. Some failures in this mode were observed when a large force was exerted along the axis of the cylindrical grip, for example when the drill was rotated. A further failure mode occurred when the fingers reached their grasping point at a higher position above the grasping surface, limiting the grasps of objects with smaller heights such as the pen, scissors, and screwdriver. This occurs because the distal joint of the fingers flex first, and the fingertips sweep through to meet at a point above the surface.

Spherical mode was used in both the YCB Gripper Assessment Benchmark (see Fig. \ref{fig:ycb_results}) and Soft Object Gripper Benchmark C (grasping crumpled clothes, see Fig. \ref{fig:soft_results_c}). This grasping mode, as expected, was successful at grasping round objects, and was also able to grasp irregularly shaped tools with a similar proficiency to the cylindrical grasp. Failure occurred during round object grasping when the objects were smaller than the caging shape of the gripper, such as the golf ball. Here, the golf ball was {grasped with the fingertips and} lifted successfully, but rolled within the gripper during rotation. {This occurred because the fingertips have relatively low friction, and the fingertip force exerted on the object has a component that acts towards the palm of the hand --- this is balanced when the hand is above the object, but during rotation the object is effectively pushed into the hand.} For larger objects, spherical grasp enabled greater form closure of the drill than cylindrical mode, meaning the drill could be rotated stably. In spherical mode, the height at which the fingertips meet is lower than in cylindrical mode. This is because the fingers are offset laterally and therefore the plane in which they flex does not pass through the centre of the gripper. {In} spherical mode, the planes intersect earlier into the flexion motion, meaning that the point at which the fingertips meet is closer to the grasping surface. This allowed the gripper to pick up the screwdriver successfully, which was not possible in cylindrical mode. Spherical mode was successful at grasping and lifting every item of clothing except the jeans in a crumpled state. Similar to the issue faced when grasping the jeans in precision mode, the weight and stiffness of the material made it difficult to grasp the jeans unless a graspable feature was already present; it was difficult to deform the jeans with the gripper to create graspable features. In the compressed crumpled state, there are a higher density of graspable features on the item of clothing, enabling grasping.

Finally, the challenging manipulation task of grasping a bunch of grapes then picking a single grape was performed as a pilot task to demonstrate the versatility of the Hydra Hand. Shown in Fig. \ref{fig:demonstration}, the bunch of grapes was caged using spherical grasping mode, then placed at a secondary location. Once placed, a single grape was picked from the bunch using the precision grasping mode with a hard coded twist-and-pull action {performed by the robot arm}. To grasp a delicate, articulate object such as a bunch of grapes is challenging for many grippers, and is only possible because of the energetically favourable flexion motion of the distal joints of the fingers, which allows the gripper to cage objects without exerting excess forces on them. Picking a single grape from the bunch is also challenging, requiring precise motion of the fingers of the gripper to not disturb other grapes in the bunch and to hold the grape firmly enough to pick it, but not so hard that it crushes the grape.


\section{Conclusions \& Future Work}\label{sec:conclusions}
\begin{figure*}[!t]
    \centering
    \vspace{0pt}
    \includegraphics[width=0.9\textwidth]{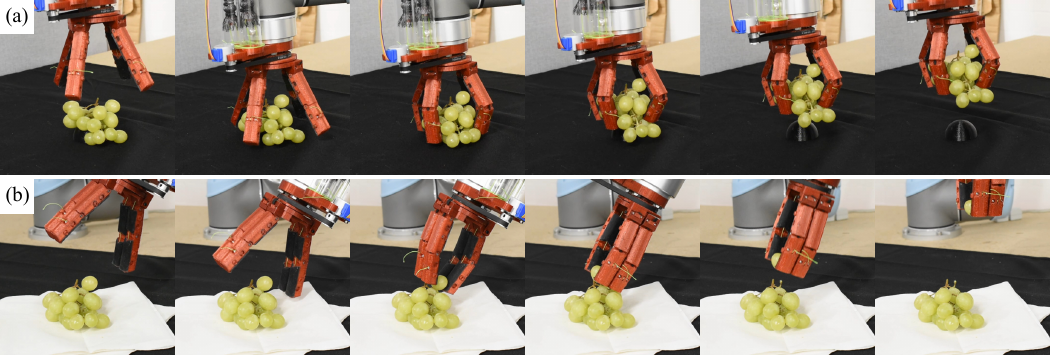}
    \vspace{-5pt}
    \caption{A demonstration of the Hydra Hand on a challenging manipulation task. (a) Using spherical mode to grasp a bunch of grapes via caging. (b) Using precision mode to pick a single grape from the bunch.}
    \label{fig:demonstration}
\end{figure*}
In this study we have presented the Hydra Hand, a gripper that is capable of switching between four-fingered underactuated compliant spherical and cylindrical grasping modes and a two-fingered rigid precision grasping mode. With only one grasping actuator, compliance and adaptability between pairs of fingers were achieved via underactuation and pressure distribution through McKibben muscles. Experimental results from the YCB Gripper Assessment and Soft Object Gripper Benchmarks demonstrate the ability of the gripper to perform its target grasps on rigid and deformable objects of varying size, shape, and weights. Results highlight a range of grasping applications from small objects and precise edge grasping to large objects and articulated object caging.

Future work will investigate actively using the rotating palm for in-hand manipulation, and exploring how the mode-switching behaviour can be used in planning grasping strategies of unknown objects. Generating large grasping forces in precision mode was difficult due to the physical locking of the distal joints leaving only the proximal joints to generate force. Further development is needed to address this.


\section*{Acknowledgements}
The authors thank Dongmyoung Lee, Barry Mulvey, and Peter Tisnikar for their meaningful feedback on this work.

\bibliographystyle{ieeetr}
\bibliography{references.bib}

\end{document}